\pdfoutput=1
\documentclass[11pt]{article}

\usepackage{acl}

\usepackage{times}
\usepackage{latexsym}

\usepackage[T1]{fontenc}
\usepackage[utf8]{inputenc}

\usepackage{microtype}

\usepackage{booktabs}
\usepackage{multirow}
\usepackage{verbatimbox}
\usepackage[ruled,vlined]{algorithm2e}

\usepackage[textsize=scriptsize]{todonotes}
\usepackage{color}

\usepackage{pifont}% http://ctan.org/pkg/pifont
\newcommand{\cmark}{\ding{51}}%
\newcommand{\xmark}{\ding{55}}%
\newcommand{\dataset}{\texttt{TempQA-WD}}

\usepackage{graphicx}
\usepackage{enumitem}

\title{A Benchmark for Generalizable and Interpretable Temporal Question Answering over Knowledge Bases}

\author{
Sumit Neelam, Udit Sharma, Hima Karanam, Shajith Ikbal, Pavan Kapanipathi\\
\bf Ibrahim Abdelaziz, Nandana Mihindukulasooriya, Young-Suk Lee, Santosh Srivastava \\
\bf Cezar Pendus, Saswati Dana, Dinesh Garg, Achille Fokoue, G P Shrivatsa Bhargav\\
\bf Dinesh Khandelwal, Srinivas Ravishankar, Sairam Gurajada, Maria Chang, \\
\bf Rosario Uceda-Sosa, Salim Roukos, Alexander Gray, Guilherme Lima\\
\bf Ryan Riegel, Francois Luus, L Venkata Subramaniam  \\
    \bf IBM Research
}

\begin{document}
\maketitle

\newcommand\tab[1][0.4cm]{\hspace*{#1}}

\begin{abstract}
 %Question Answering of Knowledge Bases has been a long standing problem in the Natural Language Processing domain. The goal of the task is to evaluate systems ability to reason and retrieve an answer from the knowledge base, given the natural language question. 
 
 Knowledge Base Question Answering (KBQA) tasks that involve complex reasoning are emerging as an important research direction. However, most existing KBQA datasets focus primarily on generic multi-hop reasoning over explicit facts, largely ignoring other reasoning types such as temporal, spatial, and taxonomic reasoning. In this paper, we present a benchmark dataset for temporal reasoning, \dataset, to encourage research in extending the present approaches to target a more challenging set of complex reasoning tasks. Specifically, our benchmark is a temporal question answering dataset with the following advantages: (a) it is based on Wikidata, which is the most frequently curated, openly available knowledge base, (b) it includes intermediate sparql queries to facilitate the evaluation of semantic parsing based approaches for KBQA, and (c) it generalizes to multiple knowledge bases: Freebase and Wikidata. %allows evaluating generalizable temporal KBQA systems on multiple knowledge bases (Freebase and Wikidata);  
 %Including the dataset, we present a modular temporal knowledge base question answering system on Wikidata as a baseline system. Overall, the baseline system shows a performance of XX on the dataset showing significant headroom for research in the temporal reasoning domain.
%  In addition to the new dataset, we present a modular, temporal KBQA system  \system~as a baseline. Overall, the baseline system shows a performance of 0.32 (F1 metric) on the dataset showing significant headroom for research in the temporal reasoning domain.
The \dataset~dataset is available at \url{https://github.com/IBM/tempqa-wd}.

\end{abstract}
\section{Introduction}
\label{sec:intro}
%Given a natural language question, the goal of Knowledge Base Question Answering (KBQA) systems is to reason and retrieve the answer based on the facts available in Knowledge Bases (KB). 
The goal of Knowledge Base Question Answering (KBQA) systems is to answer natural language questions by retrieving and reasoning over facts in Knowledge Base (KB).
While reasoning in KBQA is evolving as an important research direction, most existing datasets and research in this area have primarily focused on one-triple questions~\cite{bordes2015large} or multi-hop reasoning questions~\cite{lcquad2,berant2013} as illustrated in Table~\ref{tab:multi-hop}. 
Currently, there is a lack of approaches and datasets that address other types of complex reasoning, such as temporal and spatial reasoning.
In this paper, we focus on a specific category of questions called \textbf{temporal questions}, where answering a question requires reasoning about points and intervals in time. 
For example, to answer the question \textit{What team did Joe Hart play for before Man City?}, the system must retrieve or infer \textit{when} Joe Hart played for Man City, \textit{when} Joe Hart played for other teams, and which of the latter intervals occurred \textit{before} the former.

%\vspace{-0.2cm}
\begin{table}[htb]
\begin{small}
\centering
\begin{tabular}{|l|l|}
\toprule
\begin{tabular}[l]{@{}l@{}}
    \textbf{Category}\\
\end{tabular} 
& 
\textbf{Example} \\
\midrule

\begin{tabular}[l]{@{}l@{}}
    Single-Hop
\end{tabular} 
& 

\begin{tabular}[c]{@{}l@{}}

\emph{ Who directed Titanic Movie?}\\
{\bf SPARQL:}
select distinct ?a where \{\\
  wd:Q44578 wdt:P57 ?a\}\\

    \end{tabular} \\
\hline
\begin{tabular}[l]{@{}l@{}}
    Multi-hop\\
\end{tabular} 
& 

\begin{tabular}[c]{@{}l@{}}

\emph{ Which movie is directed by James Cameron }\\
\emph{ starring Leonardo DiCaprio?  }\\
{\bf SPARQL:}
select distinct ?a where \{\\
  ?a wdt:P57 wd:Q42574. \\
  ?a wdt:P161 wd:Q38111. \} \\
   \end{tabular} \\
\hline
\begin{tabular}[l]{@{}l@{}}
    Temporal
\end{tabular} 
& 
\begin{tabular}[c]{@{}l@{}}
\emph{What team did joe hart play for before man city?}\\
{\bf SPARQL:}
select distinct ?u  where \{\\
    wd:Q187184 p:P54 ?e1.  ?e1 ps:P54 wd:Q50602.\\
    wd:Q187184 p:P54 ?e2.  ?e2 ps:P54 ?u.\\
    ?e1 pq:P580 ?st1. ?e2 pq:P582 ?et2.\\
    filter (?et2 <= ?st1) \} \\
    order by desc (?et2) limit 1
\end{tabular} \\
\bottomrule
\end{tabular}
\caption{Examples of Single-hop, Multi-hop and Temporal reasoning questions on Wikidata.}
%\vspace{-0.2cm}
\label{tab:multi-hop}
\end{small}
\end{table}

Progress on Temporal KBQA is hindered by a lack of  datasets that can truely assess temporal reasoning capability of existing KBQA systems.
To the best of our knowledge, \textit{TempQuestions}~\cite{jia2018a}, \textit{TimeQuestions}~\cite{jia2021complex}, and \textit{CronQuestions}~\cite{saxena2021question} are the only available datasets for evaluating purely this aspect. These have, however, a number of drawbacks: (a) These contain only question-answer pairs and not their intermediate SPARQL queries which could be useful in  evaluating interpretability aspect of KBQA approaches based on semantic parsing ~\cite{yih2014semantic}; (b) unlike regular KBQA datasets~\cite{lcquad2, wikidata-benchmark, azmy-etal-2018-farewell} that can attest KBQA generality over multiple knowledge bases such as DBpedia, and Wikidata, these are suited for a single KB; (c) \textit{TempQuestions} uses Freebase~\cite{freebase} as the knowledge base, which is no longer maintained and  was officially discontinued in $2014$~\cite{freebase}.% In particular, to the best of our knowledge, there is no benchmark dataset for temporal questions with parallel annotations on multiple KBs.

%Past work on KBQA vary in terms of the core approach being used, that range from dependency parse-based approaches \cite{zou2014, sun2020b} to those involving end-to-end learning \cite{sun2020a, verga2020}. Many of these approaches use single KB as their knowledge source. Moreover, many of the datasets used for evaluation of these approaches are also single KB specific, such as Freebase or DBpedia. These factors are likely to have limited the generalizability of those approaches to other KBs \cite{dennis2018}. New datasets have recently been released with parallel annotations on multiple KBs \cite{lcquad2, swq, webqsp}, thus offering opportunity for generalizability across KBs. However, such datasets for complex questions are rare. 
 Our aim in this paper is to fill the above mentioned gaps by adapting the \textit{TempQuestions} dataset to Wikidata and by enhancing it with additional SPARQL query annotations. %Specifically, since \textit{TempQuestions} dataset contains complex temporal reasoning questions but on Freebase, we reused the natural language questions to create a temporal qa dataset  for WikiData~\cite{} as the Knowledge Base.
 %This would easily serve the purpose of building a dataset with parallel annotations on multiple KBs (in our case two different KBs, i.e.,WikiData and Freebase). 
 Having SPARQL queries for temporal dataset is crucial to refresh ground truth answers as the KB evolves.
We choose Wikidata for this dataset because it is well structured, fast evolving, and the most up-to-date KB, making it a suitable candidate for temporal KBQA.
Our resulting dataset thus contains parallel annotations (on both Wikidata and Freebase). This will help drive research towards development of generalizable approaches, i.e., those that could be easily be adaptable to multiple KBs.
In order to encourage development of interpretable, semantic parsing-based approaches, we (1) annotate the questions with SPARQL queries and answers over Wikidata, and (2) annotate a subset of this dataset with intermediate representations for the entities, relations, $\lambda$-expression along with SPARQL needed to answer the question.
%with information necessary for evaluating entity linking, relation linking, and intermediate sparql representation that are necessary components of a KBQA system.  

 %a subset of this dataset is also annotated with outputs expected at various intermediate stages of the pipeline.

%In this paper, we also describe a neuro-symbolic approach to temporal KBQA. Although, primary goal of this approach is to simply serve baseline for the dataset described above, it has design level novelty guided by the requirements of interpretability and generalizability\pavan{I dont think we evaluate this aspect so calling it out as novelty might not be the best idea}. It includes a novel question understanding framework, where an intermediate KB-agnostic logic representation of the question is obtained first, followed by mapping onto a KB-specific logic representation that further could be executed over KB as SPARQL queries.

The main contributions of this work are as follows:
%\vspace{-0.2cm}

\begin{itemize}
     
     \item A benchmark dataset, called \dataset, for building temporal-aware KBQA systems on \textit{Wikidata} with parallel annotations on \textit{Freebase}, thus encouraging the development of approaches for temporal KBQA that generalize across KBs.
    \item SPARQL queries for all questions in our benchmark, with a subset of the data also annotated with expected outputs from intermediate stages of the question answering process, i.e. entity and relation linking gold outputs. The goal here is to encourage interpretable approaches that can generate intermediate outputs matching those fine-grained annotations. 
%   \item A modular baseline approach, called \system, for temporal QA with novel question understanding framework to serve as a baseline for the benchmark dataset.
 \end{itemize}

\section{Related Work}

\begin{table*}
\centering
\resizebox{2\columnwidth}{!}{%
\begin{tabular}{l|l|l|c|c|c}
\hline
\textbf{Datasets} & \textbf{Knowledge Base} & \textbf{Emphasis} & \textbf{SPARQL} & \textbf{Intermediate} & \textbf{Templates}\\
 &  &  & \textbf{} &  \textbf{Evaluation} & \\ 
\hline  
QALD-9 & DBpedia & Multi-hop & \cmark & \cmark & \xmark \\
LC-QuAD 1.0 & DBpedia & Multi-hop & \cmark & \cmark & \cmark \\
LC-QuAD 2 & DBpedia, Wikidata & Multi-hop & \cmark & \cmark & \cmark \\
Simple Questions & Freebase, DBpedia, Wikidata & Single-hop & \cmark & \cmark & \xmark \\
Web Questions & Freebase
& Multi-hop
& \xmark
& \xmark
& \xmark \\
Web Question-SP
& Freebase, Wikidata
& Multi-hop
& \cmark
& \cmark
& \xmark \\
Complex Web Questions
& Freebase
& Multi-hop
& \xmark
& \xmark
& \xmark \\
TempQuestions
& Freebase
& Temporal
& \xmark
& \xmark
& \xmark \\
CronQuestions
& Wikidata subset
& Temporal
& \xmark
& \xmark
& \cmark \\
TimeQuestions
& Wikidata
& Temporal
& \xmark
& \xmark
& \xmark \\
\hline
\hline
\textbf{\dataset}
& \textbf{Freebase, Wikidata}
& \textbf{Temporal}
& \cmark
& \cmark
& \xmark \\
\hline
\end{tabular}
}
\caption{This table compares most of the KBQA datasets based on features relevant to the work presented in this paper. References: QALD-9~\cite{QALD_2017}, LC-QuAD 1.0~\cite{trivedi2017lc}, LC-QuAD 2.0~\cite{lcquad2}, Simple Questions~\cite{journals/corr/BordesUCW15}, Web Questions~\cite{berant2013}, Web Questions-SP~\cite{cai2013}, Complex Web Questions~\cite{talmor18compwebq}, TempQuestions \cite{jia2018a}}
\label{tab:related_work_datasets}
\end{table*}
%\vspace{-0.5cm}

% \textbf{{\bf KBQA Datasets:}}
Over the years,  many  question answering datasets have been developed for KBQA, such as Free917~\cite{cai2013}, SimpleQuestions~\cite{journals/corr/BordesUCW15}, WebQuestions~\cite{berant2013}, QALD-9~\cite{QALD_2017}, LC-QuAD 1.0~\cite{trivedi2017lc}, and LC-QuAD 2.0~\cite{lcquad2}. In Table~\ref{tab:related_work_datasets}, we compare each of these datasets across the following features: (a) underlying KB, including subsequent extensions, e.g. Wikidata~\cite{wikidata-benchmark} and DBpedia~\cite{azmy-etal-2018-farewell} based versions of SimpleQuestions, as well as the Wikidata subset of WebQSP ~\cite{webqsp}; %The underlying knowledge base: this includes the original knowledge base that the dataset was based on (e.g. Freebase for SimpleQuestion) and also the efforts to create parallel datasets that work on multiple KBs (e.g. Wikidata~\cite{wikidata-benchmark} and DBpedia~\cite{azmy-etal-2018-farewell} based versions of SimpleQuestions). 
%Similarly, WebQSP has a subset mapped to Wikidata~\cite{webqsp}. 
%(b) Reasoning types: the different reasoning types the dataset is attesting; e.g. single-hop, multi-hop or temporal reasoning. %As Table~\ref{tab:related_work_datasets} shows, most datasets focus of single-hop and multi-hop reasoning;
(b) reasoning types that are emphasized in the dataset;
(c) availability of SPARQL queries, entities, and relationships for intermediate evaluation; and (d) the use of question templates, which can often generate noisy, unnatural questions.
%Template-based datasets provide an option to generate a lot of data with sometimes significant noise in comparison to non-template based that are carefully curated. 
As Table~\ref{tab:related_work_datasets} shows, our dataset is distinguished from prior work in its emphasis on temporal reasoning, its application to both Freebase and Wikidata, and its annotation of intermediate representations and SPARQL queries. 

%These datasets focus on single hop and multi-hop questions ignoring complex reasoning questions. While~\cite{QALD_2017} are natural questions with SPARQL queries manually annotated,~\cite{lcquad2, trivedi2017lc} are template based datasets that allows creating of large question-sparql-answer triplets facilitating training of data hungry models for KBQA~\cite{}. 

%Not all datasets  

%Many of these are used in building semantic parsing capabilities to get the logic form closer to Freebase. ~\cite{bao-etal-2016-constraint} introduced another benchmark called ComplexQuestions on Freebase to test the multi-constraint questions. 

%Closest to our benchmark, TempQuestions~\cite{jia2018a} is constructed by gathering questions which have temporal signals from the Free917, WebQuestions, and ComplexQuestions datasets, resulting in 1272 temporal questions. As mentioned in Section~\ref{sec:intro}, we reuse the TempQuestions dataset by mapping the answers to Wikidata and annotating them with SPARQL queries that can be run on Wikidata.
The most relevant KBQA dataset to our work is TempQuestions~\cite{jia2018a}, upon which we base \dataset, as  described in Section \ref{sec:dataset}.
CronQuestions~\cite{saxena2021question} is another dataset where emphasis is on temporal reasoning. However, this dataset also provides a custom KB derived from Wikidata which acts as a source of truth for answering the questions provided as part of the dataset.
\section{Dataset}
\label{sec:dataset}
 %\textit{TempQuestions} \cite{jia2018a}, a benchmark dataset for temporal KBQA is created by extracting temporal questions from three different KBQA datasets: 
 \textit{TempQuestions} \cite{jia2018a} was the first KBQA dataset intended to focus specifically on temporal reasoning. It consists of temporal questions from three different KBQA datasets with answers from Freebase: \textit{Free917} \cite{cai2013}, \textit{WebQuestions} \cite{berant2013} and \textit{ComplexQuestions} \cite{bao2016}. %This was the first attempt at emphasizing and encouraging temporal reasoning for KBQA. 
 %In order to have a temporal QA dataset that can be:  (a) generalizable benchmark with parallel annotations on multiple knowledge bases; (b) up-to-date knowledge base such as Wikidata; and (c) intermediate evaluation with SPARQL queries, we decided to adapt \textit{TempQuestions} to Wikidata. 
 We adapt \textit{TempQuestions} to Wikidata to create a temporal QA dataset that has three desirable properties. First, in identifying answers in Wikidata, we create a generalizable benchmark that has parallel annotations on two KBs. Second, we take advantage of Wikidata's evolving, up-to-date knowledge. Lastly, we enhance \textit{TempQuestions} with SPARQL, entity, and relation annotations so that we may evaluate intermediate outputs of KBQA systems.

 %However, the gold answers for these datasets are based on \texit{Freebase} which was officially discontinued from year $2014$ \cite{freebase}. 
 
 %It is desirable to have a benchmark dataset, especially for temporal questions, on a KB that is up to date, and not frozen in time. \textit{Wikidata} \cite{wikidata} offers a good option, as it is also better structured, apart from being more up-to-date and evolving.

%Instead of creating a new dataset with a fresh set of questions, we decided to adapt \textit{TempQuestions} dataset \cite{jia2018a} to \textit{WikiData}, as this also would easily serve the purpose of building a dataset with parallel annotations on multiple KBs (in our case two different KBs, i.e., \textit{WikiData} and \textit{Freebase}). As explained in the previous section, we believe this likely will play a role in helping drive research towards development of generalizable approaches, i.e., those that could be easily adaptable to multiple KBs. \pavan{We need to emphasize the TempQuestions does not have SPARQL queries where as having an intermediate representation would allow 1. Allow KG independent research }

%There has been attempts to transfer KBQA questions to \textit{WikiData} by WebQSP-WD~\cite{} and SimpleQuestions-WD~\cite{}. They directly map the answers of Freebase to their corresponding entity in Wikidata. However, this has multiple challenges: 
Two previous attempts at transferring Freebase-QA questions to \textit{Wikidata} are WebQSP-WD~\cite{webqsp} and SimpleQuestions-WD(SWQ-WD)~\cite{diefenbach2017wdaqua}. SWQ-WD is single triple questions and in WebQSP-WD  only answers are directly mapped to corresponding entities in Wikidata. However, as stated ~\cite{webqsp}, one challenge is that not all \textit{Freebase} answers can be directly mapped to entities in \textit{Wikidata}. For example, the Freebase answer annotation for the question "When did Moscow burn?"  is ``1812 Fire of Mosco'', despite the year being entangled with the event itself. In contrast, Wikidata explicitly represents the year of this event, with an entity for ``Fire in Moscow'' and an associated year of ``1812". Thus, a direct mapping between the two answers is not possible, as it would amount to a false equivalence between ``1812 Fire of Mosco'' and ``1812''.

In order to address such issues, %we opted to take on the tedious task of manually writing SPARQL queries, 
we enlisted a team of annotators to manually create and verify SPARQL queries, %which allows for answers'  verification and updates as and when there are periodic updates on Wikidata. 
ensuring not only that the SPARQL formulation was correct, but that the answers accurately reflected the required answer type (as in the ``Fire in Moscow'' example above) and the evolving knowledge in Wikidata (as the Freebase answers from the original dataset may be outdated). 
Having SPARQL queries also facilitates intermediate evaluation of the approaches that use semantic parsing to directly generate the query or the query graph%~\cite{ijcai2020}
, increasing interpretability and performance in some cases~\cite{webqsp}. 
%At the minimum, manually written SPARQL queries on \textit{WikiData} are needed for each question, in order to verify if the answer is present or not, and note down if available. 
%Writing SPARQL queries need manual search within the KB for matching entities/predicates and navigating around to estimate the reasoning required to be encoded in the SPARQL. Some questions could fall-out in this process depending upon non-availability of the answers or unavailability of the required reasoning paths. 
% However, given that we are considering temporal questions, presence of necessary time interval information about events on Wikidata is critical, which is not always the case.

%Next, we give a brief overview of how \textit{Wikidata} is organized, before explaining the dataset creation process.
Next, we give a brief overview of how \textit{Wikidata} is organized, as some of its representational conventions impact our dataset creation process.

\subsection{Wikidata}
\begin{figure}[t]
  \centering
  \includegraphics[scale=0.36]{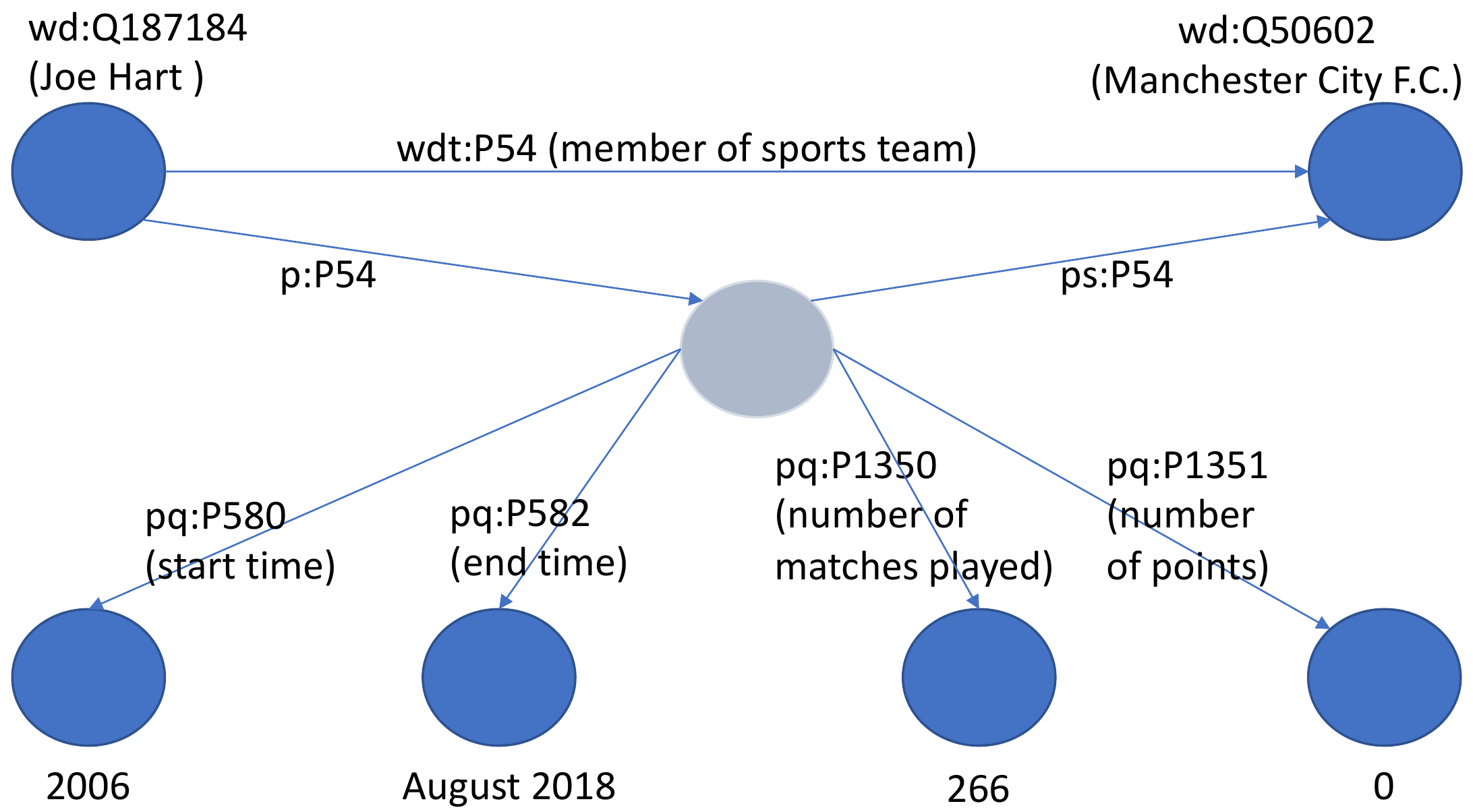}
  \caption{An illustration of Wikidata Reification.}
  %\vspace{-0.2cm}
  \label{fig:temporal_reasoning_example} 
\end{figure}

Wikidata\footnote{https://www.wikidata.org/} is a free knowledge base that is actively updated and maintained. It has 93.65 million data items as of May, 2021 and continuously growing each day with permission for volunteers to edit and add new data items. We chose Wikidata as our knowledge base as it has many temporal facts with appropriate knowledge representation encoded. %embedded with the statements in the form of reified statements, making it an interesting KB for testing temporal question answering. 
It supports reification of statements (triples) to add additional metadata with qualifiers such as start date, end date, point in time, location etc. 
%to capture valid time interval of the statement, location where the event has occurred or any other details that are relevant to the system. 
Figure~\ref{fig:temporal_reasoning_example} shows an example of reified temporal information assocaited to entities Joe Hart and Manchester City. 
With such representation and the availability of up-to-date information, Wikidata makes it a good choice to build benchmark datasets to test different kinds of reasoning including temporal reasoning.

%\vspace{-0.2cm}

\subsection{Dataset Details}
\label{sec.dataset_details}

\begin{table}[htb]
\begin{small}
\centering
\begin{tabular}{c|c|c|c}
\hline
\hline
Dataset & Size & Answer & Add'l \\
& & & Details\\
\hline
\hline
\textit{TempQuestions} & 1271 & \textit{Freebase} & Set-A \\
\textit{(Freebase)}                      &      & Answers & \\ 
\hline
\textit{Wikidata} & 839 & \textit{Wikidata} & Set-A + \\
\textit{Benchmark}   &      & Answers & Set-B \\ 
\hline
\textit{A Subset} & 175 & \textit{Wikidata} & Set-A + \\
\textit{of Wikidata} & & Answers & Set-B + \\
\textit{Benchmark}                &     & & Set-C \\
\hline
\hline
\end{tabular}
\caption{Benchmark dataset details. \textit{TempQuestions} denote original dataset with \textit{Freebase} answers \cite{jia2018a}. The remaining two, i.e., \textit{testset} and \textit{devset}, are part of the benchmark dataset we created with \textit{WikiData} answers. In column titled \textit{Add'l Details}: Set-A denote additional annotations that came along with the original \textit{TempQuestions} dataset, \{temporal signal, question type, data source\}. Set-B denote \{\textit{Wikidata} SPARQL query, question complexity category\}. Set-C denote ground truths of \{AMR, $\lambda$-expression, \textit{Wikidata} entities, \textit{Wikidata} relation, \textit{Wikidata}-specific $\lambda$-expression\} }
\label{tab:benchmark}
\end{small}
\end{table}

\begin{figure*}[t!]
  %\centering
  \includegraphics[keepaspectratio, width=1.0\textwidth]{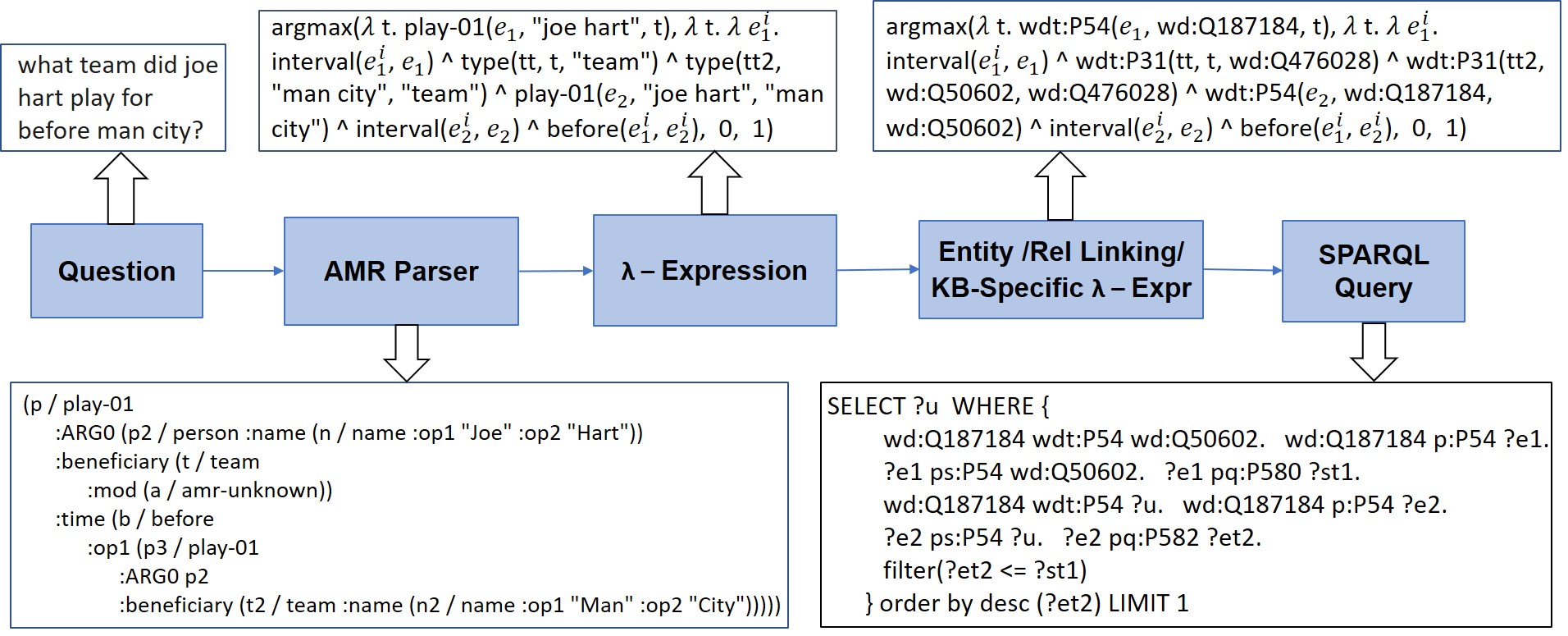}
  \caption{Additional annotations example}   
  \label{fig:example}
\end{figure*}
Table~\ref{tab:benchmark} gives details of our new benchmark dataset. We took all the questions from \textit{TempQuestions} dataset (of size $1271$) and chose a subset for which we could find \textit{Wikidata} answers. This subset has 839 questions that constitute our new dataset, \dataset. We annotated this set with their corresponding \textit{Wikidata} SPARQL queries and the derived answers. We also retained the \textit{Freebase} answers from the original \textit{TempQuestions} dataset effectively creating  parallel answers from two distinct KBs. Additionally, we  added question complexity label to each question, according to question complexity categorization described in Section~\ref{sec:category}. 

Within this dataset, we also chose a smaller subset (of size $175$) for more detailed annotations. Details of those additional annotations can be found in the Caption of Table~\ref{tab:benchmark} and an  %\colorbox{yellow}{an} 
illustration of such annotations for a sample question can be found in Figure~\ref{fig:example}. 
 %{\color{red}can be found in the caption} of Table~\ref{tab:benchmark}. \pavan{Can you please add more information on what the detailed annotations are with example?}. 
The goal of these additional annotations is to encourage improved interpretability of the temporal KBQA systems, i.e., to evaluate accuracy of outputs expected at intermediate stages of the system. 
%We believe these additional annotations are potentially useful to evaluate modular approaches for KBQA. 
Modular approaches such as NSQA\cite{pavan2020} and SYGMA\cite{neelam2021sygma} are more appropriate for KBQA, given difficulty in gathering large amount of training data to build end-to-end systems. In fact, we use SYGMA\cite{neelam2021sygma} as our baseline system to evaluate this benchmark dataset. Such approaches stack together, in a pipeline, various modules built elsewhere with minimal adaptation.
In such a setup, it is important to evaluate outputs at intermediate stages of the pipeline. Additional annotations include, ground truths for intermediate meaning representations such as AMR, $\lambda$-expression of the question and KB-specific $\lambda$-expressions, and mention-of/KB-specific entities and relations.

\subsubsection{Question Complexity Categorization}
\label{sec:category}
\begin{table}[htb]
\begin{small}
\centering

\begin{tabular}{|l|l|}
\toprule
\begin{tabular}[l]{@{}l@{}}
    \textbf{Category}\\
    \textbf{(dev/Test)}
\end{tabular} 
& 
\textbf{Example} \\
\midrule
% \hline

\begin{tabular}[l]{@{}l@{}}
    Simple\\
    (92/471)
\end{tabular} 
& 

\begin{tabular}[c]{@{}l@{}}

{\bf When was Titanic movie released?}\\
SELECT ?a WHERE \{ \\
    wd:Q44578 wdt:P577 ?a \}\\
    
   \end{tabular} \\
\hline
\begin{tabular}[l]{@{}l@{}}
    Medium\\
    (71/154)
\end{tabular} 
& 

\begin{tabular}[c]{@{}l@{}}

{\bf who was the US president during} \\
{\bf cold war?   }\\
SELECT DISTINCT ?a WHERE \{\\
  ?a wdt:P39 wd:Q11696. \tab
  ?a p:P39 ?e1.  \\
  ?e1 ps:P39 wd:Q11696. \\
  ?e1 pq:P580 ?st1. \tab
  ?e1 pq:P582 ?et1.  \\
  wd:Q8683 wdt:P580 ?st2. \\
  wd:Q8683 wdt:P582 ?et2. \\
FILTER (?st1 <= ?et2 \&\& ?st2 <= ?et1)\}
 \\

  \end{tabular} \\
\hline
\begin{tabular}[l]{@{}l@{}}
    Complex\\
    (12/39)
\end{tabular} 
& 
\begin{tabular}[c]{@{}l@{}}

{\bf who was president of the us when }\\
{\bf douglas bravo was a teenager? } \\
SELECT DISTINCT ?a WHERE \{\\
  ?a p:P39 ?e. \tab
  ?e ps:P39 wd:Q11696.\\
  ?e pq:P580 ?st1.\tab
  ?e pq:P582 ?et1.\\
  wd:Q4095606 wdt:P569 ?x.\\
  bind ((?x + “P13Y”$^{\wedge\wedge}$xsd:duration) as ?st2)\\
  bind ((?x + “P19Y”$^{\wedge\wedge}$xsd:duration) as ?et2)\\
  FILTER (?st1<=?et2 \&\& ?st2<=?et1) \}\\
\end{tabular} \\

\bottomrule
\end{tabular}
\caption{Examples of Simple, Medium, and Complex temporal reasoning question in the dataset.}
\label{tab:complexity_examples}
\end{small}
\end{table}

For evaluations in Section~\ref{sec:eval}, we divided dataset \dataset~of size $839$ into two parts,
\dataset-test of size $664$ and \dataset-dev of size $175$. In fact, \dataset-dev is that part of the dataset with fine-grained annotations. In this dataset, we also labeled questions with complexity category based on the complexity of the question in terms of temporal reasoning required to answer. Table~\ref{tab:complexity_examples} shows the examples for each category of complexity defined below.
\vspace{0.1cm}
%\begin{itemize}
%\item 

\noindent \textbf{1) Simple:} Questions that involve one temporal event and need no temporal reasoning to derive the answer. For example, questions involving simple retrieval of a temporal fact or simple retrieval of other answer types using a temporal fact. 

%\item 
\noindent \textbf{2) Medium:} Questions that involve two temporal events and need temporal reasoning (such as overlap/before/after) using time intervals of those events. We also include those questions that involve single temporal event but need additional non-temporal reasoning.

%\item 
\noindent \textbf{3) Complex:} Questions that involve two or more temporal events, need one temporal reasoning and also need an additional temporal or non-temporal reasoning like teenager or spatial or class hierarchy.

%Table~\ref{tab:complexity_examples} shows number of dev and test questions for each category along with an example question and its SPARQL query.
%\end{itemize}

%Data and annotation details:
%* TempQA - 1272
%      - Ques + GT Freebase Ans + temporal signal + question type + data source
%* Our test - 500 
%      -  Ques + GT Wikidata SPARQL + GT WikiData Ans + Complexity Categories + metadata from orig. data.
%* Our dev - 175
%      - similar info as above + GT AMR + GT Lambda Expression + GT WikiData entities/relations + GT KB-Specific Lambda

%Complexity categorization of question with respect to Wikidata.
%1) Simple - Single temporal event with no temporal reasoning required - simply retrieving a temporal fact / retrieving answer based on a single temporal fact.
%2) Medium - Two temporal events and a temporal reasoning across events like overlap, before after case / single temporal event together with additional non-temporal reasoning.
%3) Complex - Two or more temporal events  - with additional temporal events like teenager / including additional spatial reasoning or class hierarchy reasoning.
\subsection{Lambda Calculus}
\label{sec:lambda}
%Although deriving KB-specific query of a NL question is the end goal of KBQA system, it is useful to also look the intermediate meaning representations of the questions. 
%For example, modular approaches, like our baseline described in Section \ref{sec:baseline}, would generate such intermediate representations. We use $\lambda$-expression for this purpose.
%For this, we use $\lambda$-calculus based logical representation of the question, i..e, $\lambda$-expression of the question.

%This is modular approaches are to likely need to attractive alternatives \cite{pavan2020}, given difficulty in building end-to-end systems that need large amount of training data.
%Modular approaches stack together, in a pipeline, various modules that are built elsewhere and typically need only minimal adaptation.
%In this context, for our dataset, it is useful to not only have the annotations for the end goal of KB-specific queries, but also have annotations for the intermediate outputs of different modules. These include modules that output intermediate meaning representations. 

In this section, we give a brief description of $\lambda$-calculus, since we use $\lambda$-expressions to logically represent the semantics of questions. $\lambda$-calculus, by definition, is considered the smallest universal programming language that expresses any computable function.
%In our pipeline Lambda calculus plays vital role in representing intermediate logical form of question. By definition, $\lambda$ calculus is smallest universal programming language as it facilitates to express any computable function using its formalism. It mainly focuses on applying transformation rules to express computable function. 
In particular, we have adopted \textit{Typed} $\lambda$-\textit{Calculus} presented in \cite{zettlemoyer2012learning}. 
%In our representation we have used Constants, Logical Connectives similarly. However, we have introduced some new temporal quantifiers and instance variable in lambda expression to avoid function nesting.
In addition to the constants and logical connectives, we introduced some new temporal functions and instance variables to avoid function nesting.  
For example, consider the following question and its corresponding logical form:  
\begin{center}
\begin{tabular}{l c}
 Question: & when was Barack Obama born?\\ 
 Logical Form: & \begin{tabular}{l l}$\lambda$\textit{t. born(b,``Barack Obama")} $\wedge$\\\textit{interval(t, b)} \end{tabular}\\  
\end{tabular}
\end{center}
Here, \textit{b} is instance variable for event \textit{born(b, ``Barack Obama")} and \textit{interval(t, b)} finds time for the event denoted by \textit{b}. Variable \textit{t} is unknown which is marked as $\lambda$ variable.

%Consider the example question text and its logical form. In logical form, \textit{b} is instance variable for \textit{born(b,``Barac Obama")}. \textit{interval(t, b)} finds time for event \textit{b}. Variable \textit{t} is unknown which is marked as lambda variable.
%\subsubsection{Temporal Functions:}
\noindent
\textbf{Temporal Functions:} We introduce \textit{interval, overlap, before, after, teenager, year}; where \textit{interval} gets time interval associated with event and \textit{overlap, before, after} are used to compare temporal events. \textit{teenager} gets teenager age interval for a person, and \textit{year} return year of a date.
% We have adopted following $\lambda$ calculus grammar from \cite{Rojas15}:
% \begin{center}
% \begin{tabular}{l l l}
%  <expression>&:=&\begin{tabular}{l l}<name> | <function> |\\<application> \end{tabular}\\ 
%  <function>&:=&$\lambda$<name>.<expression>\\  
%  <application>&:=&<expression><expression>\\
% \end{tabular}
% \end{center}
% Notice that, \textit{expression} is main concept in $\lambda$ calculus, however our usage is limited to function only. We have skipped detailed description of grammar as it is widely known and available.

%$\lambda$t. born(b,``Barac Obama")$\wedge$time(t, b)

%\input{text/data/analysis}

% \input{text/approach}
\section{Evaluation}
\label{sec:eval}
%In this section, we describe evaluation of \system~ on \dataset. In addition, to also illustrate the effectiveness of \system~ for questions beyond temporal in nature, we evaluated it on two other Wikidata benchmark data sets.

\subsection{Evaluation Setup}
We use SYGMA\cite{neelam2021sygma} to the generate the baseline results for \dataset~dataset. The baseline system is tuned with dev set of TempQA-WD , and evaluated on the test dataset.
% We implemented our pipeline using a Flow Compiler%\footnote{https://github.com/IBM/flow-compiler}
% ~\cite{chakravarti-etal-2019-cfo} stitching individual modules exposed as gRPC services. We defined the ANTLR grammar to define $\lambda$-expressions.
% %, that includes rules to capture basic predicates, logical connectives (like $and, or, not$), basic temporal operators (like $argmin, argmax, min, max$), temporal operators (like {\it interval, overlap, before, after, teenager, now, age}) and so on. 
% KB-Specific $\lambda$-expression to SPARQL module is implemented in Java using Apache Jena %\footnote{https://jena.apache.org/} 
% SPARQL modules, to create SPARQL objects and generate the final SPARQL query, to run on public Wikidata end point. %\footnote{https://query.wikidata.org/} 
% Rest of the module are implemented in Python and are exposed as gRPC services.  
% %We use GERBIL%\footnote{https://github.com/dice-group/gerbil}
% %~\cite{DBLP:journals/semweb/UsbeckRHCHNDU19} to compute performance metrics using the gold answers and the system generated answers from pipeline.
% \input{text/eval/datasets}
\subsection{Metrics}
%We use the standard macro precision, macro recall and F1 measures normally used to benchmark the KBQA systems. All the numbers produced are using GERBIL by providing the gold answers and system generated answers to it.

We use GERBIL%\footnote{https://github.com/dice-group/gerbil}
~\cite{DBLP:journals/semweb/UsbeckRHCHNDU19} to compute performance metrics from the pairs of gold answers and system generated answers from the pipeline.
We use standard performance metrics typically used for KBQA systems, namely macro precision, macro recall and F1.
\subsection{Results \& Discussion}

Table~\ref{table:f1} shows performance of SYGMA along with the 
% s described in Section \ref{sec:data}.
%We selected these datasets as they have more natural questions.
%{\achille{We should also report results from other QA systems operating on Wikidata on the three datasets or say explicitly that no such comparable systems exist }}
performance of TEQUILLA%\footnote{https://tequila.mpi-inf.mpg.de/}
~\cite{jia2018b} (that uses Freebase as its underlying KB) on \dataset ~test set.
% To our knowledge, ours is the first system reporting temporal QA performance on Wikidata.
The accuracy numbers show that there is room for improvement and good scope for research on temporal QA (and complex QA in general) on Wikidata.
Table \ref{table:category} gives detailed report of performance for different categories of question complexity.
Performance of TEQUILLA reflects how well Accu and Quint (the QA systems it used) are adapted to Freebase.
This in addition to the fact that majority of the questions fall under simple questions category (471 simple vs 39 complex as shown in Table~\ref{tab:complexity_examples}), it is able to achieve better accuracy on Freebase. Given relatively less focus of research on temporal QA on Wikidata so far, we believe our benchmark dataset would help accelerate more research and datasets in the future.

%We ran TempQA-WD test set on TEQUILLA\footnote{https://tequila.mpi-inf.mpg.de/}~\cite{jia2018b} that uses Freebase as its underlying KB. %We Tuned our system using 175 dev questions from TempQA-WD and SWQ-WD datasets and ran the same pipeline to test on WebQSP-WD dataset as well. 
%We achieve better performance as~\cite{webqsp} (0.25) on the same test set without any tuning.
%of baseline and TEQUILLA systems on the on further categorization present in the dataset. Note that dataset is categorized purely based on the reasoning requirements and not based on individual component complexity like entity extraction linking or relationship linking etc. So, the numbers presented may not truly reflect the reasoning complexity depending on how the individual modules perform.

To gain more insights on the performance, we also did an ablation study of SYGMA using \dataset~dev set where impact of individual module on overall performance is evaluated. Table~\ref{table:ablation} shows the results.
%We took intermediate ground truths available in that dataset and fed them directly as input %to appropriate modules in the pipeline, 
%to assess the impact of performance of different modules.
% We did similar study also with SWQ-WD dataset, however it has ground truth annotations only for AMR and entities. 
%In the table, row \textit{Machine} refer to full end-to-end performance without use of any ground truths. Rows GT-$Module$ are for the use of intermediate ground truths. 
For example GT-$AMR$ refers to the case where ground truth AMR is fed directly into $\lambda$-module. The table shows large jump in accuracy (in both the datasets) when fed with ground truth entities (GT-EL) and ground truth relations (GT-RL). This points to the need for improved entity linking and relation linking on Wikidata.

\begin{table}[!t]
\small
\begin{tabular}{|l|llll|}
\hline
    System&  Dataset  & Precision & Recall & F1 \\
\hline
 SYGMA &TempQA-WD &  0.32  &    0.34     &  0.32       \\
% \system &SWQ-WD &    0.54       &   0.43     &  0.44       \\
%  &WebQSP-WD &       0.32     &   0.29     &    0.28      \\
\midrule
 TEQUILLA&TempQA-FB  &    0.54      &    0.48   &   0.48      \\
\hline
\end{tabular}
\caption{Performance of SYGMA }
\label{table:f1}
\end{table}

\begin{table}[hbt]
\small
\begin{center}
\begin{tabular}{|l|llll|}
\hline
 System&Category & Precision & Recall & F1 \\
 \hline
 &Simple  & 0.39 & 0.37 & 0.38               \\
 SYGMA &Medium  & 0.16 & 0.13 & 0.13                \\
 &Complex  & 0.38 & 0.38 & 0.38               \\
\midrule
        & Simple  &    0.59     &    0.53  &   0.53      \\
TEQUILA& Medium &0.46 & 0.39& 0.40\\
    & Complex & 0.20&0.20 & 0.18\\
\hline

 \end{tabular}
\caption{Category wise Performance}
\label{table:category}
\end{center}
\end{table}

%\begin{table*}[htb]
%\small
%\begin{tabular}{|llllllllll|}
%\hline
%    System&  Dataset  &Metric& %Machine & GT-AMR & GT-$\lambda$ %& GT-EL &GT-RL& GT-KB$\lambda$& %GT-SPARQL \\
% \hline
% & &Precision  &0.49 & 0.51&0.53 %& 0.61&0.92 & 0.93& 1.0    \\
%OurSystem&TempQA-WD &Recall  & 50.47&0.49 &0.52 & 0.60& 50.92&0.92 & 1.0    \\
%& &F1  & 0.47& 0.50&0.52 & 0.60& %0.92& 0.92&1.0     \\
% \midrule
% & &Precision  & 0.47&0.58 & - & %0.82& - & - & 1.0    \\
%OurSystem &SWQ-WD &Recall  &0.47 %&0.59 & - &0.82 & - & - & 1.0    %\\
%& &F1  &0.47 & 0.58& - & 0.82& - %& - &1.0     \\
%\hline
%\end{tabular}
%\caption{Ablation Study on the %Dev Set }
%\label{table:ablation}
%\end{table*}

\begin{table}[!t]
\small
\begin{center}
\begin{tabular}{|l|ccc|}
%\begin{tabular}{|llllllllll|}
\hline
& \multicolumn{3}{|c|}{TempQA-WD} \\
\hline
 & Precision & Recall & F1  \\
\hline
NO GT & 0.47 & 0.50  & 0.47 \\
GT-AMR & 0.50 & 0.51  & 0.50 \\
GT-$\lambda$ & 0.52 & 0.53 & 0.52\\
GT-EL & 0.60 & 0.62 & 0.60\\
GT-RL & 0.92 & 0.93 & 0.92\\
GT-KB-$\lambda$ & 0.93 & 0.93 & 0.93 \\
GT-SPARQL & 1.0 & 1.0 & 1.0 \\
\hline
\end{tabular}
\caption{Ablation Study on \dataset}
\label{table:ablation}
\end{center}
\end{table}

\section{Conclusion}

In this paper, we introduced a new benchmark dataset \dataset~for temporal KBQA on \textit{Wikidata}. Adapted from existing \textit{TempQuestions} dataset, this dataset has parallel answer annotations on two KBs. %thus provide opportunity to evaluate generalizability of the temporal KBQA approaches. In addition to the answers annotations, 
A subset of this dataset is also annotated with output expected at intermediate stages of modular pipeline. %This offers opportunity to evaluate interpretability of the approaches. 
%Apart from benchmark dataset, 
% We also described a temporal KBQA system \system~that serves as a baseline on the dataset. 
%This system is designed developed with goals of generalizability and interpretability. 
Future extenstions include %extension of the dataset also to KBs beyond Wikidata and Freebase, extension of the 
improvement of the baseline approach for generalizable temporal QA.

\bibliography{anthology,custom}
\bibliographystyle{acl_natbib}

\end{document}